\documentclass[letterpaper, 10 pt, conference]{ieeeconf}  

\IEEEoverridecommandlockouts                              

\usepackage{caption}
\usepackage{subcaption}
\usepackage{gensymb}
\usepackage{amsmath}
\usepackage{amsfonts}
\usepackage{xcolor}
\usepackage{soul}
\usepackage{hyperref}
\usepackage{setspace}
\usepackage{booktabs}
\usepackage{algorithm2e}
\hypersetup{colorlinks,linkcolor={red},citecolor={blue},urlcolor={red}}  
\RestyleAlgo{ruled}
\SetKwComment{Comment}{/* }{ */}
\usepackage{cite}
\usepackage{graphicx}

\title{\LARGE \bf
Reconfigurable Robot Identification from Motion Data
}

\author{Yuhang Hu* \quad Yunzhe Wang \quad Ruibo Liu \quad Zhou Shen \quad Hod Lipson\\
Columbia University
\thanks{This work was supported in part by the US National Science Foundation (NSF) AI Institute for Dynamical Systems (DynamicsAI.org), grant 2112085
        {\tt\small yuhang.hu@columbia.edu}. For more information: https://github.com/H-Y-H-Y-H/meta\_self\_modeling\_id }%
}

\begin{document}

\maketitle
\thispagestyle{empty}
\pagestyle{empty}

\begin{abstract}

Integrating Large Language Models (LLMs) and Vision-Language Models (VLMs) with robotic systems enables robots to process and understand complex natural language instructions and visual information. However, a fundamental challenge remains: for robots to fully capitalize on these advancements, they must have a deep understanding of their physical embodiment. The gap between AI models' cognitive capabilities and the understanding of physical embodiment leads to the following question: Can a robot autonomously understand and adapt to its physical form and functionalities through interaction with its environment? This question underscores the transition towards developing self-modeling robots without reliance on external sensory or pre-programmed knowledge about their structure. Here, we propose a meta-self-modeling that can deduce robot morphology through proprioception—the robot's internal sense of its body's position and movement. Our study introduces a 12-DoF reconfigurable legged robot, accompanied by a diverse dataset of 200k unique configurations, to systematically investigate the relationship between robotic motion and robot morphology. Utilizing a deep neural network model comprising a robot signature encoder and a configuration decoder, we demonstrate the capability of our system to accurately predict robot configurations from proprioceptive signals. This research contributes to the field of robotic self-modeling, aiming to enhance robot's understanding of their physical embodiment and adaptability in real-world scenarios. 

\end{abstract}

\section{INTRODUCTION}
The development of artificial general intelligence (AGI) capable of controlling robots in the real world necessitates a deep understanding of the robot's physical embodiment. Recent research has focused on integrating Large Language Models (LLM) or Vision-Language Models (VLM) with robots to enhance their capabilities\cite{ahn2022can, cao2023ground,yu2023l3mvn,ding2023task}. However, most approaches rely on human prompts to guide robots in completing specific tasks. LLM and VLM can extract information in task scenarios for robots, allowing high-level controllers to complete decision-making. However, these methods still require pre-programming of some behaviors, which limits the robot's capabilities. For example, a robot can derive the position of a mug on the table from visual information, but how to grab the mug still needs to rely on the underlying human program. Therefore, in order to create a robot that effectively incorporates the capability of LLM or LVM into its control strategy as human common sense, the model must understand the robot body in the physical world.

Traditionally, robots have been programmed with predefined models that describe their kinematics, dynamics, and physical structure (often represented as URDF - Universal Robot Description Format files). Relying solely on predefined models can limit a robot's adaptability and resilience. These models are typically static and do not change. Therefore, they might not account for wear and tear, modifications to the robot's body, or entirely new environments. To overcome these limitations, there's a growing interest in developing self-modeling robots\cite{bongard2006resilient,yang2018grand,cully2015robots}. These robots can understand and update their models through their own experiences. This capability is crucial for life-long learning, enabling robots to adapt to changes in their physical structure or environment.

\begin{figure}[t]
  \centering
  \includegraphics[width=0.48\textwidth]{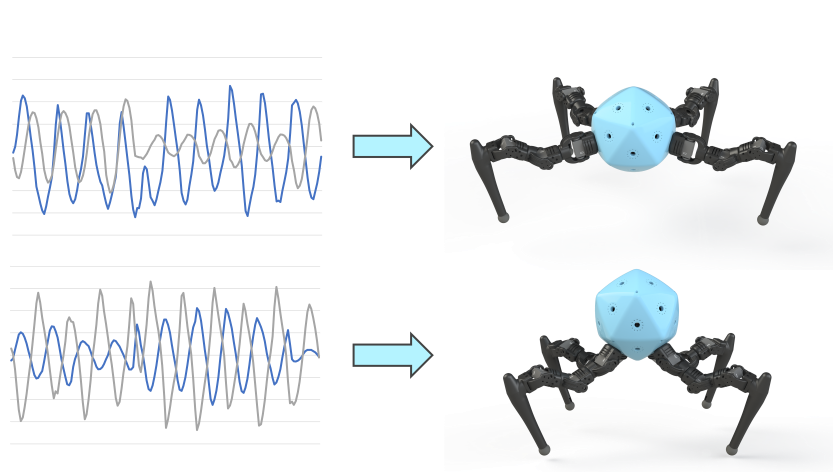}
  \caption{\textbf{Configuration prediction from motion data.} To what degree is it possible to reconstruct the topology of a robot from its motion dynamics alone? (concept illustration only) For more insights into the motivation behind our research, we invite readers to view the supplementary videos.}
  \vspace{-15pt}
  \label{fig:concept}
\end{figure}

To enable robots to comprehend the ontology of the physical world, relying on external sensors to understand robot ontology might not be universally effective. This is due to the variability in robots' operational environments and the diversity in the calibration and configuration of external sensors. For example, a robot equipped with two cameras with different configurations would capture disparate images, leading to variations in perception. In contrast, focusing on proprioception could offer a more consistent and effective means for these models to understand a robot's kinematics, dynamics, and morphology. Proprioception provides direct insights into a robot's internal state and dynamics, which are less affected by external environmental factors. This approach might streamline the process of integrating robots with advanced cognitive models by offering a more standardized basis for understanding robotic systems.

In robotics, the intricate relationship between morphology and motion dynamics plays a pivotal role. For example, is it possible to tell if a robot has two legs or four legs simply by observing time series data of its body accelerations and joint angles? While it is clear that the dynamics of a slithering snake are different than those of a bipedal walker, the question is to what degree can this information alone be used to recover the complete morphology of a robot (Fig.\ref{fig:concept}). This paper delves into the possibility of reconstructing a robot's topological structure based solely on its motion dynamics data from proprioception.
\begin{figure*}[!h]
  \centering
  \includegraphics[width=0.9\textwidth]{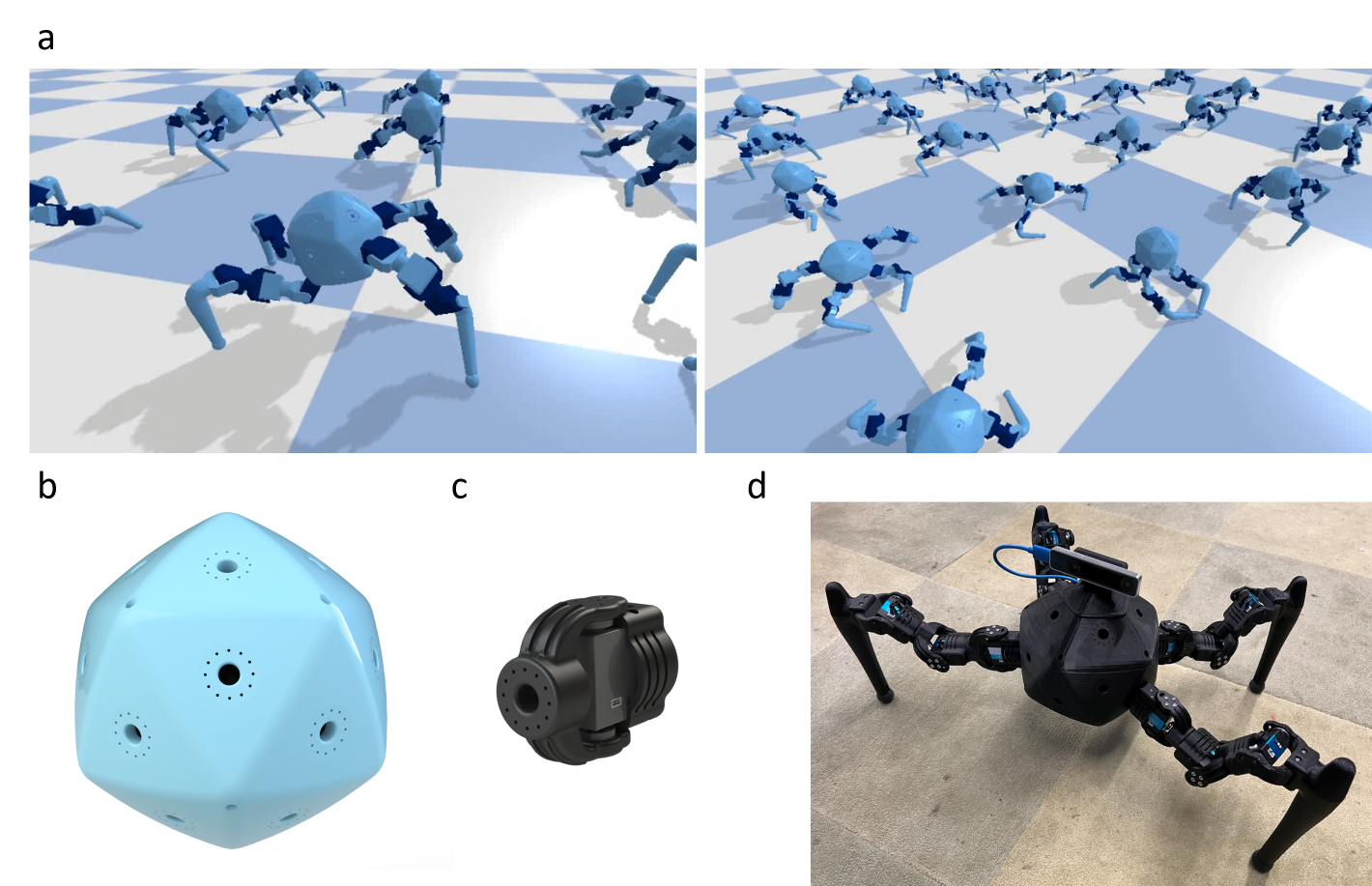}
  \caption{A comprehensive view of the reconfigurable robots used in our work. (a) Overview of reconfigurable robots exhibiting a variety of configurations in the simulation environment. (b) Detailed view of the robot's main body, a geometrically precise icosahedron with 20 uniform faces designed for versatile connection to joint modules. (c) Close-up of a single joint module, equipped with an individual motor, demonstrating its potential for connection at 12 distinct angles to allow for a broad range of movement and reconfiguration. (d) Fully-assembled assembled physical robot in the real world.}
  \vspace{-10pt}
  \label{fig:rcfg}
\end{figure*}

The ability to create a long-term body self-image through movement is natural for humans \cite{proske2012proprioceptive, o1995proprioception,hart2014robot}: we can perceive the position, orientation and motion of our body parts without seeing and thinking, thanks to proprioceptors located within muscles, tendons, skin and joints \cite{jankowska1992interneuronal}.  Such spatial self-awareness is also essential for robots to anticipate outcomes of motor action without trying them out in physical reality. Past research has tried to replicate such capability through computational robot self-model\cite{kwiatkowski2022origins,boulgouris2005gait}. 

The strategy employed in this study also bears a resemblance to Biometric Motion Identification in human studies, where individuals can be distinguished based on motion capture data obtained from kinetic videos and 3D skeletal sequences. By mirroring this approach, we aim to imbue our robotic self-model with the capability to learn and predict new and unseen configurations \cite{han2017space,munsell2012person}. This could aid with detecting change to the body of existing robots (e.g. due to damage or failure), or speed up the modeling of new robots by automatically deciphering their topology.

In this work, we proposed a learning framework to identify individual robots from their proprioceptive data.  We designed a 12-DoF reconfigurable legged robot, creating a diverse dataset of 200k unique robotic configurations, each distinctly represented via a signature code as shown in Fig.\ref{fig:rcfg}. By using a single reconfigurable legged robot platform, we create a controlled environment that can systematically explore a wide range of configurations. We introduce a meta-self-modeling, a Multiclass-Multioutput Robot Morphology Classifier, which allows a robot to form an understanding of its body morphology from a limited set of self-movement data. Once trained, the meta-self-model can predict unseen robot configurations by just observing proprioceptive signals. Through this methodology, our meta-self-model has the capacity to comprehend and interpret robot dynamics, marking a stride forward in the realm of robotic self-modeling. The main contributions of this work are as follows:

The motivation of our work is not limited to the identification and prediction of robot configurations but extends into the realm of applying our model's latent space for broader applications. If our framework, based on proprioception data, can accurately classify robot configurations, this implies that the model's latent space possesses the ability to differentiate between different robots based on proprioceptive information alone. Such a richly informed latent space opens up new avenues for applying these insights to other tasks, such as designing adaptive controllers, predicting dynamics with precision, or even visualizing the morphologies of robots in ways previously unattainable. In essence, the foundational understanding of robot morphologies and dynamics, facilitated by our model's insights into proprioception data, can serve as a critical bridge for LLM and VLM to perform general real-world physical interactions. The main contributions of this work are as follows:

1. We present a meta-self-modeling approach that aims to learn shared dynamics across diverse reconfigurable robot morphologies. This approach allows the system to understand multiple robot dynamics and predict specific configurations through proprioceptive data, marking progress in robotic self-modeling.

2. The development of a 12-DoF reconfigurable legged robot. This design allows for a high degree of versatility in physical configurations, enabling the study of the relationship between robotic motions and robot configurations within a single platform.

3. We open-source a diverse dataset of 200k unique configurations from a 12-DoF reconfigurable legged robot with the same icosahedron body. This dataset includes robot URDF files, initial joint positions, and the hardware CAD design. The entire robot can be fabricated by FDM 3D-printing for easy reproducibility and further research, enabling other researchers to print and assemble the robots.

\section{Related Work}
\textbf{Self-modeling Robots.}
The concept of self-modeling permeates various disciplines, from human cognition and animal behavior to robotic systems \cite{gallup1982self, rochat2003five,bongard2006resilient, chen2022fully}. Essentially, it involves an agent—be it biological or artificial—constructing an internal representation of its physical properties. In the robotic field, self-modeling is one of the data-driven control systems providing a computational representation of the different aspects of robots, including morphology, kinematics, and dynamics \cite{dearden2005learning}, as well as other facets like sensing, actuation, control, and planning\cite{wolpert1998internal}. A well-trained Self-model can be implemented as a predictive model for Model Predictive Control\cite{hu2022egocentric}. It segregates the model of the robot from a model of its environment and task. The robot itself is relatively consistent across different tasks and environments, so isolating the self-model for reuse simplifies adaptation in varying scenarios, even resilience from the damaged body, thus facilitating transfer learning betwen robots \cite{cully2015robots,chen2022fully,kwiatkowski2019task}.

\textbf{Robotic System Identification.}
 Bongard and Lipson \cite{1492385} introduced a coevolutionary algorithm for inferring hidden nonlinear systems. Wu and Movellan \cite{6385977} proposed the Semi-Parametric Gaussian Processes (SGP), merging the advantages of parametric and non-parametric system identification approaches for underactuated robotics. Wenhao Yu et al. \cite{yu2017preparing} developed a Universal Policy (UP) and Online System Identification (OSI) function to adapt to unknown dynamic models. Bruder et al. \cite{8793766} introduced a system identification technique using Koopman operator theory in the domain of soft robotics. 
 The landscape of learning-based controllers for robotic systems is also vast and rapidly evolving. Researchers present learning universal policies understanding the relationship between morphology and control, and employing meta-learning for improving learning efficiency\cite{rakelly2019efficient,kurin2020my,huang2020one}. Gupta et al.'s "MetaMorph"\cite{gupta2022metamorph}  and Trabucco et al.'s "AnyMorph"\cite{trabucco2022anymorph} both focus on universal controllers, with the former utilizing Transformers and the latter emphasizing the learning of transferable policies by inferring agent morphology
 Our work extends these foundations by harnessing shared dynamics among diverse robot morphologies for system identification. 
 
\textbf{Reconfigurable Robots.}
Due to the capability of Reconfigurable robots that can reconstruct their morphology, they have found significant application in space exploration, where compactness is a crucial factor, and other fields requiring versatile adaptability \cite{yim2007modular, castano2001representing}. Optimizing these reconfigurable robots often involves searching for optimal configurations for specific tasks. Prior studies have designed and explored reconfigurable manipulators, where the robot's structure can be adjusted to suit different tasks \cite{ceccarelli2004multi,yun2020modman}. Furthermore, the complexity of reconfigurable systems has been pushed even further with the underactuated robots, offering greater adaptability and control in complex dynamic environments\cite{ha2018computational,kim2017snapbot,ha2016task}. Although these systems provide diverse morphology, their usage in learning shared dynamics across different robot configurations has been limited. The complex, high-dimensional configuration space presented by reconfigurable robots presents an ideal opportunity for training a meta-self-model that learns the shareable essence of robot dynamics.

\section{Method}

\subsection{Robot Configuration Name}
\label{sec:cfg_name}
\begin{figure}[!h]
  \centering
  \includegraphics[width=0.48\textwidth]{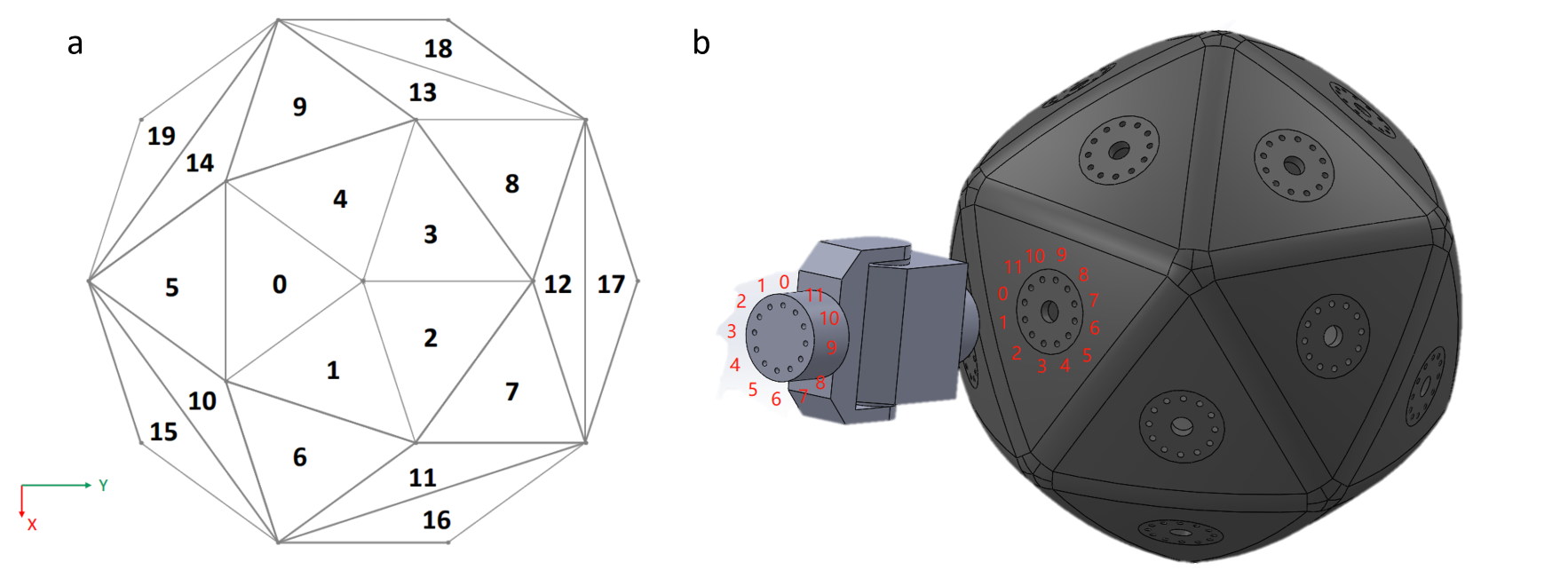}
  \caption{Coding method of the icosahedron body and the angle of each link. a) Integer vector coding for the icosahedron body. b)Integer vector coding for the angle of the twelve links. Each face of the icosahedron body is sequentially numbered in a counterclockwise direction from top to bottom. The connected joints allow rotation angles, divided into 12 segments with a 30-degree separation, also numbered in a counterclockwise manner.}
  \vspace{-10pt}
  \label{fig:cfg_name}
\end{figure}

To study a robot family with similar configurations, we designed a 12-DoF quadruped robot with URDF descriptions to be loaded in physics simulation engines and assembled in the real world. Its legs can be attached to any four faces of an icosahedron body, and each leg consists of three links where the connection point between links can be rotated to 12 different angles with 30 degrees separation counterclockwise. This gives a family of robots with a total number of $C^{20}_4\cdot (4\cdot 3)^{12}\approx 4.32\times 10^{16}$ possible configurations depending on how we assemble the robot. To uniquely describe each symmetric-leg robot, we designed an integer vector coding  $\mathbf{y} \in \mathbb{Z}^{12}$ such that $0\leq y_f \leq 19$ for $f\in \{0, 4\}$, indicating the two faces one one side with legs, and that $0\leq y_l\leq 11$ for $l \in \{ 1,2,3,5,6,7 \}$, indicating the angles of the six links (Fig.\ref{fig:cfg_name}). When calculating errors using the L1 distance, this encoding ensures that adjacent configurations (like "0" and "11") have a difference of 1. Therefore, the maximum distance achievable is 6. For example, when the true label is "0" but the predicted label is "6". We implemented a script to generate the corresponding URDF file, which was given a configuration for simulation.

\subsection{Data Collection}
\label{sec:data}

Our data collection schema consists of two phases: robot generation and dynamic data collection. We randomly generated 200K robots with different configuration coding, where each code maps to unique URDF files describing the robot's morphology. All generated robots are symmetric and can stand when loaded into the simulation.

\begin{algorithm}[h]
\caption{Robot Generation}
\label{algo:robot_gen}
\SetKwFunction{RandInt}{RandInt}
\SetKwFunction{URDF}{URDF}
\SetKwFunction{collide}{collide}
\SetKwFunction{slip}{slip}
\KwData{Left faces indices $I_f$, Number of robots $N$, Configuration mirror function $m$}
\KwResult{Robot Dataset $\mathcal{D}_r$ as a mapping from config code to URDF}
$\mathcal{D}_r \gets \emptyset$\;
\While{$|\mathcal{D}_r| < N$} {
    $C_{J} \gets \RandInt{range=12, n=6}$\;
    $C_{L} \gets \{C_{L} \subseteq I_f \mid card(C_{L})=2$\}\;
    $C \gets [C_{L}; m(C_L); C_J; m(C_J)]$\;
    $R \gets \URDF{C}$\;
    \eIf{\collide{$R$} or \slip{$R$}}{
        \textbf{continue}  
    }{
        $\mathcal{D}_r \gets \mathcal{D}_r \cup \{C: R\}$\;
    }
}
\end{algorithm}
\vspace{-10pt}

We randomly generated 163k robots for the reconfigurable legged robot dataset with different configuration codes as described in algorithm \ref{algo:robot_gen}. We generated the corresponding URDF description file for each configuration coding and loaded them in the PyBullet Physics Engine for validation. We filtered out robots with self-collision and those that would slip over when loaded into the simulation. Slipping over is triggered when the robot's body roll or pitch value is greater than $\pi/2$ while it moves. For simplicity, we constrained the robot structure in two ways: 1) only 12 out of the 20 faces will be chosen to attach legs, where all of them are located in the middle and bottom layers of the icosahedron; 2) All robots are symmetric, meaning that given half of the configuration code (8 values) on one side of the robot, we are guaranteed to determine the other half through a mirror function $m$. This simplification drastically reduces the total number of possible configurations to $2\cdot C^{6}_2\cdot (2\times 3)^{12}\approx 8.96\times 10^{7}$. Despite the simplification, it still remains challenging to predict the structure of these robots due to the vast array of potential configurations.

\subsection{Model Architecture}
We proposed a deep neural networks model that consists of two components: a robot signature encoder and a configuration decoder. The encoder handles both channel-wise and temporal dependencies of the collected state sequences, extracting a latent robot morphological representation. The decoder has seven classification heads; each is a single fully connected layer and decodes the latent representation to the leg-face pattern and joint orientations. 

\begin{figure*}
  \centering
  \includegraphics[width=0.95\textwidth]{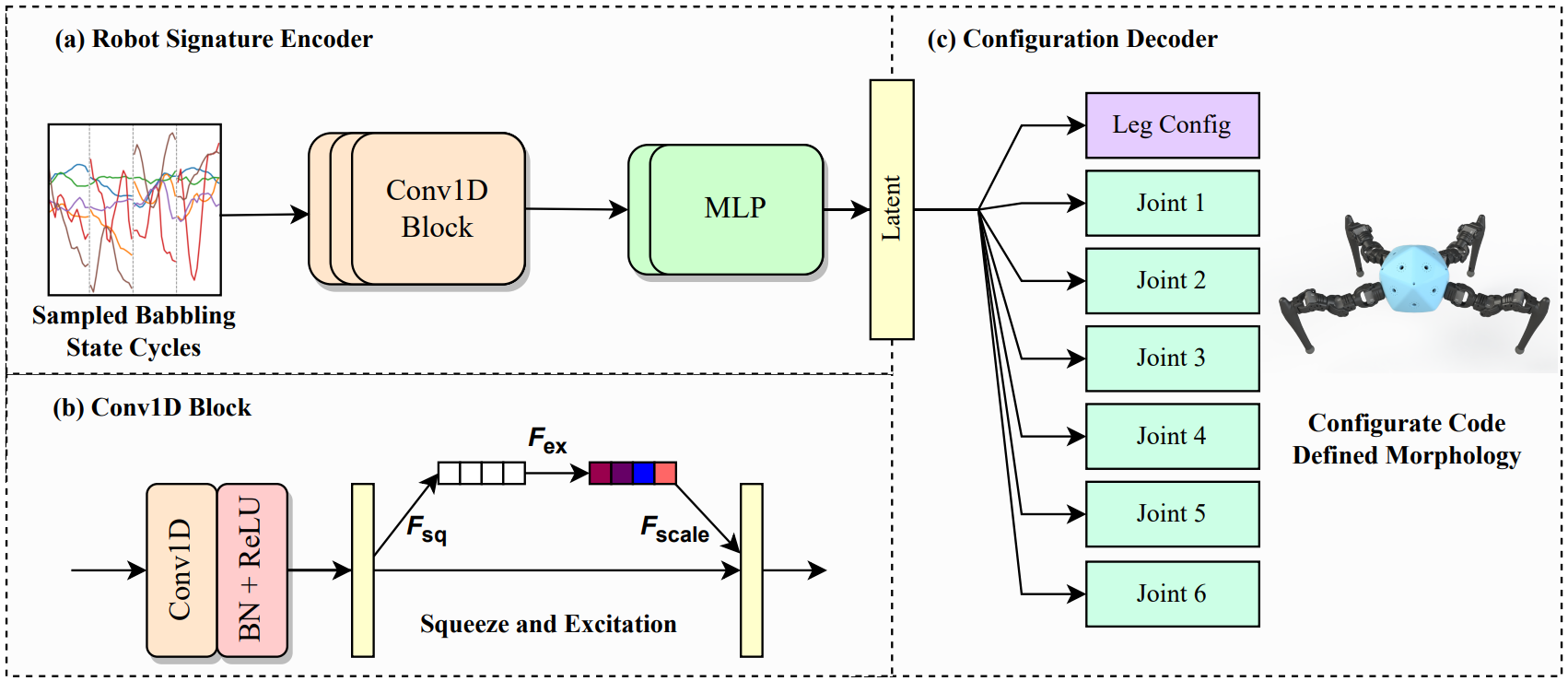}
  \caption{The model architecture of the classifier. The robot signature encoder (a) employs three 1D convolution blocks for channel-wise dependency. Each convolution block (b) employed the squeeze and excitation operation. The spatial features are then encoded through a 2-layer MLP network into a latent vector. Lastly, the latent vector was decoded by seven prediction heads for leg positions and six joints on one side of the robot, where each one is a single fully connected layer. The predicted indices for leg positions and joint angles are selected by taking $argmax$ over each head's output. }
  \vspace{-10pt}
  \label{fig:diagram}
\end{figure*}

\paragraph{Robot Signature Encoder} Given the collected state-sequences of a robot, we aim to predict its configuration code to differentiate individual robots by observing its movement. We name this process ``robot signature encoding,'' similar to biometric motion identification, where a person can be recognized based on his/her motion capture sequence. Fig.\ref{fig:diagram}(a) illustrates the process where dynamic data obtained from sampled robot babbling is introduced into the Robot Signature Encoder. We leverage 1D convolution layers with squeeze-and-excitation blocks \cite{hu2018squeeze} for capturing channel-wise dependencies as illustrated in Fig.\ref{fig:diagram}(b). The encoded features are concatenated and fed through Multi-Layer Perceptron (MLP) layers as a latent vector $\mathbf{z}=\Phi_{\mathrm{ENC}}(\mathbf{x})\in \mathbb{R}^d$ before decoding the configuration code.

\paragraph{Configuration Decoder}
To predict the configuration code vector $\mathbf{y}$, we leveraged a multi-output decoder $\Phi_{\mathrm{DEC}}$ with seven prediction heads as present in Fig.\ref{fig:diagram}(c) that decodes $\mathbf{y}$ from the latent space $\mathbf{z}$. One head classifies the leg position pattern with 30 choices, and the other six heads classify the orientation pattern for the six legs on one side of the robot; each has 12 choices. Due to the symmetry constraint, the other half of the robot legs can be completely determined. Therefore, a 7-heads classifier is sufficient for our setup:

$$\Phi_{\mathrm{DEC}}:=(\Phi_{\mathrm{LEG}}, \Phi_{\mathrm{JNT}}^1, \Phi_{\mathrm{JNT}}^2, \Phi_{\mathrm{JNT}}^3, \Phi_{\mathrm{JNT}}^4, \Phi_{\mathrm{JNT}}^5, \Phi_{\mathrm{JNT}}^6)$$

\paragraph{Objective Function}

We applied the Categorical Cross-Entroy loss function for each classification head.

\begin{equation}
    \mathcal{L}_{head} = - y \log \frac{\exp(\hat{y})}{\sum_{c=1}^C \exp(\hat{y}_i)} 
\end{equation}

Where $\hat{y}$ is the network output logits, $y$ is the ground truth label, and $C$ is the number of classes, where $C=30$ if \verb|head=leg| and $C=12$ if \verb|head=joint|. The total loss across all seven heads is aggregated by taking a weighted sum controlled by a ratio hyper-parameter $\lambda$ due to the varying difficulties between classifying legs and joints.

\begin{equation}
    \mathcal{L}_{total} = \lambda \mathcal{L}_{leg} + (1-\lambda) \frac{1}{6} \sum_{i=1}^6 \mathcal{L}_{joint}^{(i)} 
\end{equation}

\paragraph{Input Data Sampling}
We adopted a sampling approach for the input state sequences during training. During data collection, we collected the motion dynamics of the robot with a size of $16\times100\times30$. Rather than taking all collected trajectories as input, we sampled 10 continuing trajectories as input $\mathbf{x}$. Each trajectory is a vector of $16\times30$, representing the 30 robot state data over 16 steps. For each robot configuration training instance, we forward the input data with a size of $16\times10\times30$ through the model to get the predicted logits $\hat{\mathbf{y}}$ and updated the gradient based on the configuration label $\mathbf{y}$ and the loss function $\mathcal{L}_{total}$ with a ratio hyper-parameter $\lambda=0.75$ determined empirically. We split the data with an 8:2 ratio for training and validation.

Our model parameters are optimized using the PyTorch \cite{paszke2019pytorch} deep-learning framework and adaptively tuned the model hyperparameter using the HYPERBAND \cite{li2017hyperband} algorithm through the Weight-and-Bias \cite{wandb} package. We employed the Adam Optimizer \cite{kingma2014adam} with learning rate $3\text{e-}4$ and weight decay $1\text{e-}5$. We used the ReLU activation function, Batch Normalization \cite{ioffe2015batch}, and a Dropout \cite{srivastava2014dropout} probability of 0.3 for all convolution layers and the first MLP layer. We also applied gradient clipping with value 1 for the LSTM module to avoid gradient exploding. The training was done over one Nvidia RTX2080 GPU for 800 epochs with batch size 128, taking 38 hours in total.

\section{Experiments}

\subsection{Dynamic Data Collection} 
For each valid robot configuration, we collected their state sequence for ten trials, each containing ten steps of motor babbling cycles. During random movement, the robot might also fall over before finishing 6 steps. If that happens, the trial would be aborted and rerun as shown in algorithm \ref{algo:motor_bab}.

For each robot, we collected its dynamic state (position, orientation, joint angles) sequences while performing random motor babbling actions defined by the parametric sine-gait function in equation \ref{eq:sine_gait}. $A$ is the amplitude parameter and $\phi$ is the phase-shift parameter. $i$ indexes the three joints of the same leg (ordered as inner, middle, and outer joints), and $j$ indexes the four legs (ordered as right-hind, right-front, left-front, and left-hind legs). $t$ is the current timestep, and $\tau=16$ is a predetermined period constant indicating the number of sub-steps within a cycle. The action $a_{ij}$ indicates the targeted angle of the joint $i$ in leg $j$ at timestep $t$. During motor babbling, $A$, $\phi$, and robot initial joints position $\theta_{ij}$ are chosen uniformly at random with values normalized to the motor action space. Meanwhile, robot state sequences are collected at each timestep.

\begin{equation} \label{eq:sine_gait}
    a_{ij} = A_i \cdot \sin \left(  \frac{t \ \mathrm{mod}\ \tau}{\tau }\cdot 2\pi + \phi_j  \right) + \theta_{ij}
\end{equation}

At timestep $t$, the robot's state $S_t\in \mathbb{R}^{18}$, consists of the position and Euler-angles of the icosahedron body at the center of mass $[x,y,z, \psi, \theta, \phi]$ as well as the angles of the 12 joints. The dynamic state data over an entire babbling cycle can thus be seen as a multinomial time series of 18 channels with sequence length $\tau$. In terms of actions, the robot's subsequent action at timestep $t+1$  is represented as $A_{t+1} \in \mathbb{R}^{12}$. Each dynamic motion data incorporates 30 parameters: 18 from state data $S_t$ and 12 from next actions $A_{t+1}$.

\begin{algorithm}
\caption{Dynamic Data Collection}
\SetKwFunction{obs}{state}
\SetKwFunction{step}{next\_step}
\SetKwFunction{Rand}{Rand}
\KwData{Robot Dataset $\mathcal{D}_r$, Number of steps $N$, Number of trajectories $T$, Sine-gait action function $a$.}
\KwResult{Dynamic Dataset $\mathcal{D}_k$}
\ForEach{$\{C:R\} \in \mathcal{D}_r$}{
    $\mathcal{D}_k^{(C)} \gets \emptyset$\;
    \While{$|\mathcal{D}_k^{(C)}|<T$}{
        $\mathcal{D}_t \gets \emptyset$\;
        \For{$n=\{1,...,N\} $}{
            $\theta\gets\Rand{n=10}$\;
            $R$.\step($a(\theta)$)\;
            \If{\slip{$R$}}{
                \textbf{break}
            }
            $\mathcal{D}_t \gets \mathcal{D}_t \cup \{\obs{$R$}\}$\;
        }
        $\mathcal{D}_k^{(C)} \gets \mathcal{D}_k^{(C)} \cup \mathcal{D}_t$\;
    }
}
\label{algo:motor_bab}
\end{algorithm}

\subsection{Baselines}
To evaluate the effectiveness of our approach, we designed three baselines. BL-LSTM baseline is a variation of our proposed method (OM-Conv1). The motivation behind using this baseline is to evaluate the temporal learning capacity of LSTM cells when employed within our framework\cite{hochreiter1997long}. LSTMs, with their inherent capability of learning long-term dependencies, could be a promising alternative for handling sequences. In this variation, we replace the Conv1D Block (Fig.\ref{fig:diagram}) with two layers of Long Short-Term Memory (LSTM) cells. It allows us to discern whether a dedicated temporal model can perform comparably or better than our Conv1D block.

With the BL-MLP baseline, we aim to explore the efficiency of simple feed-forward neural networks on our task. By replacing the Conv1D blocks with fully connected (dense) layers, the architecture essentially turns into a multi-layer perceptron (MLP) model. This variation is critical as it measures how well non-sequential models perform side-by-side with our proposed method and other sequential methods. 

Recognizing that a conventional Inertial Measurement Unit (IMU) struggles to accurately capture the center of mass (CoM) positional data (x,y,z), we incorporated a baseline to simulate such a real-world scenario. By training the model without CoM location data as input, we seek to understand how important this data is for effective modeling and prediction within our framework.

\subsection{Evaluation Metrics}
During the training and evaluation phases, we observed three key metrics.
\textbf{Leg-Acc:} Prediction accuracy for the leg position configuration in percentage. \textbf{Jnt-Acc-Avg:} Average prediction accuracy over 6 joints. \textbf{Tot-Acc:} Prediction accuracy over legs and all 6 joints. We used a $L_1$ distance error function during the evaluation between the prediction and the ground truth for a more intuitive view. When calculating the errors, the L1 distance between configurations "0" and "11" is 1, so the largest error is 6.

\subsection{Quantitative Evaluations}
During the model training phase, each model variation was trained ten times, each time with different hyperparameters. After these runs, we selected the best model with the minimum loss for the evaluation. Our evaluation utilized a test dataset comprising 40k robots. The process involved inputting the state sequence and predicting the robot configuration name. The $l_1$ distance error of each prediction head is detailed in Table \ref{tab:head_acc}.

\begin{table}[h!]
    
    \centering
    \caption{Accuracy of predicting joint configuration}\label{tab:head_acc}
    \resizebox{\columnwidth}{!}{
    \begin{tabular}{cccccccc}
    \hline
        \textbf{} & \textbf{} & \textbf{Jnt 1} & \textbf{Jnt 2} & \textbf{Jnt 3} & \textbf{Jnt 4} & \textbf{Jnt 5} & \textbf{Jnt 6 } \\ \hline
        \textbf{Acc Mean} & OM & \textbf{0.749} & \textbf{0.743} & \textbf{0.746} & \textbf{0.746} & \textbf{0.752} & \textbf{0.744}  \\ 
        \textbf{} & OM-rm\_xyz & 0.592 & 0.586 & 0.588 & 0.590 & 0.590 & 0.587  \\ 
        \textbf{} & BL-LSTM & 0.629 & 0.629 & 0.625 & 0.623 & 0.631 & 0.633  \\ 
        \textbf{} & BL-MLP & 0.267 & 0.268 & 0.265 & 0.263 & 0.267 & 0.269  \\  \hline
        \textbf{Err-Dist Mean} & OM & \textbf{0.057} & \textbf{0.059} & \textbf{0.059} & \textbf{0.058} & \textbf{0.057} & \textbf{0.058}  \\ 
        \textbf{} & OM-rm\_xyz & 0.104 & 0.107 & 0.107 & 0.105 & 0.106 & 0.105  \\ 
        \textbf{} & BL-LSTM & 0.087 & 0.088 & 0.090 & 0.089 & 0.087 & 0.087  \\ 
        \textbf{} & BL-MLP & 0.271 & 0.273 & 0.272 & 0.274 & 0.274 & 0.273  \\  \hline
        \textbf{Err-Dist Std.} & OM & \textbf{0.129} & \textbf{0.131} & \textbf{0.134} & \textbf{0.128} & \textbf{0.128} & \textbf{0.129}  \\ 
        \textbf{} & OM-rm\_xyz & 0.172 & 0.177 & 0.177 & 0.172 & 0.176 & 0.174  \\ 
        \textbf{} & BL-LSTM & 0.152 & 0.154 & 0.158 & 0.155 & 0.153 & 0.153  \\ 
        \textbf{} & BL-MLP & 0.268 & 0.268 & 0.268 & 0.268 & 0.270 & 0.269  \\ \hline
    \end{tabular}
}
\vspace{-5pt}
\end{table}

\begin{figure}
  \centering
  \includegraphics[width=0.48\textwidth]{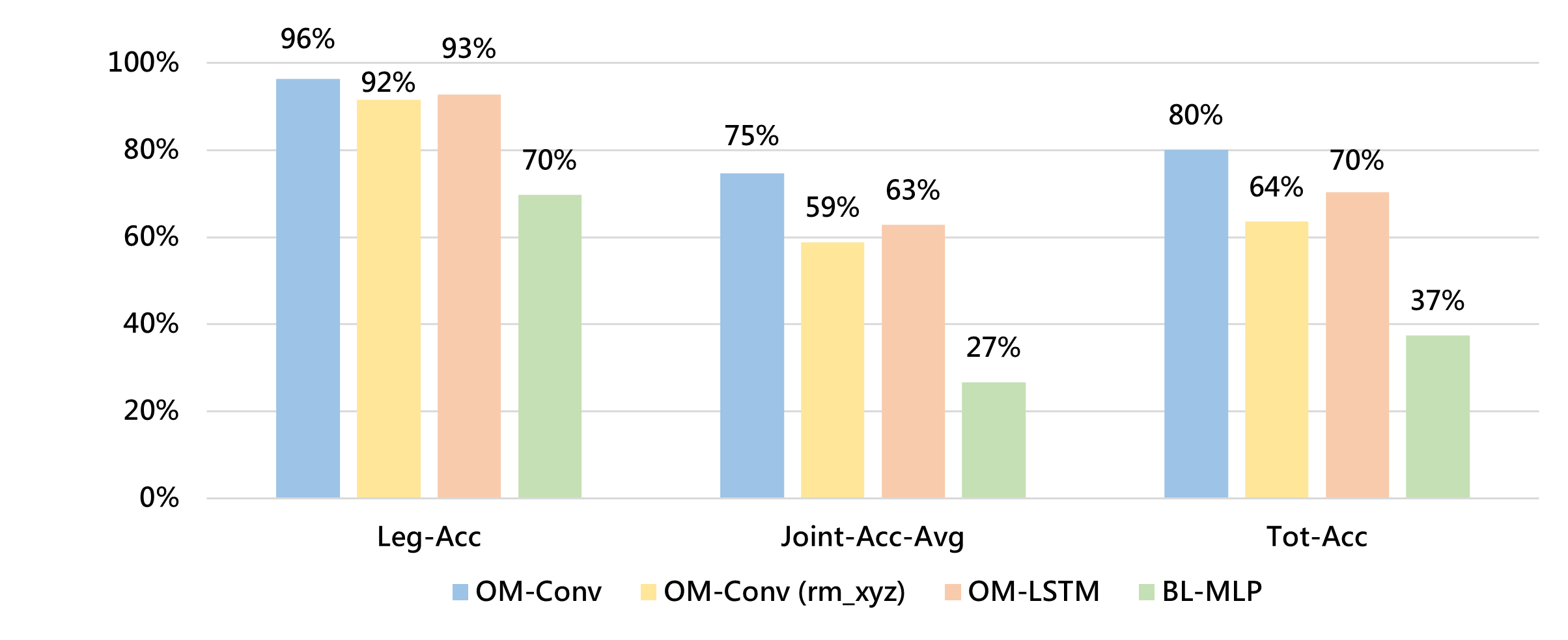}
  \caption{Performance Comparisons. This figure presents bar plots comparing the leg accuracy, average joint accuracy, and total accuracy of our proposed method against the three baselines. The results show that our method outperforms the baselines across all metrics.}
  \vspace{-5pt}
  \label{fig:quan}
\end{figure}

Inside the table, \textbf{Acc:} Prediction accuracy for joint angle configuration in percentage. Jnt1 and Jnt4 are inner joints, Jnt2 and Jnt5 are middle joints, Jnt3 and Jnt6 are outer joints. \textbf{Err-Dist:} The $l_1$ distance between the predicted and real configurations integer vector. 
We altered our model's architecture using various modules to assess the encoder module's performance within the meta-self-model and different input data. Our results in the accompanying Fig.\ref{fig:quan} and Tab.\ref{tab:head_acc} show that our method with a convolutional module in the encoder can provide satisfied performance, predicting leg configurations with 96\% accuracy. 

\begin{figure}
  \centering
  \includegraphics[width=0.48\textwidth]{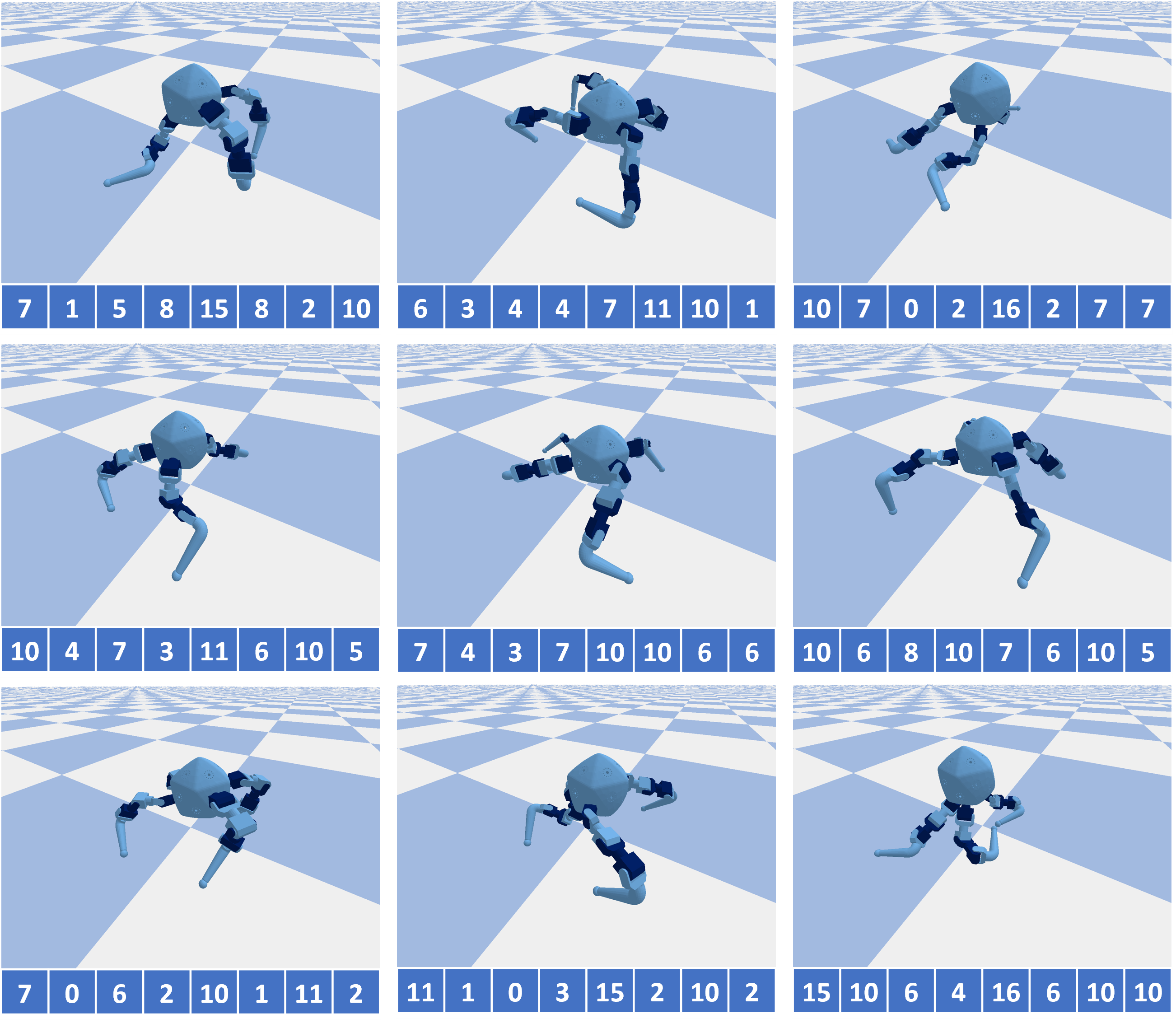}
  \caption{Test Dataset Samples Visualization. A selection of nine robots was randomly chosen from the test evaluation dataset. Our model successfully identified both the leg configurations and all 6 joint positions accurately, yielding a 100\% accuracy rate for these samples. The robot configuration name is denoted under each image.}
  \vspace{-10pt}
  \label{fig:sim_experiment}
\end{figure}

In Figure\ref{fig:sim_experiment}, we show some robot pictures that the meta-self-model prediction achieved a 100\% hit rate for both robot joint and leg positions. The displayed configurations further emphasize the model can consistently identify a wide spectrum of robot morphologies.

\subsection{Real-world Experiment Results}
\begin{figure}[h]
  \centering
  \includegraphics[width=0.48\textwidth]{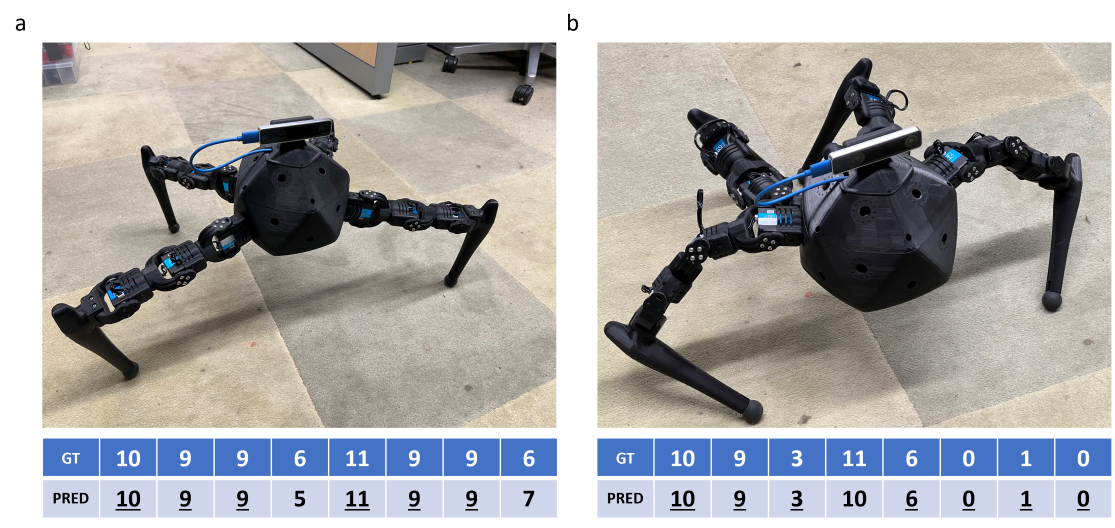}
  \caption{Real-World Testing of Reconfigurable Robots with Their Corresponding Configuration Name. a) Image of two tested robots, showing their unique configuration as utilized in the real-world experiment. The configuration names for both robots are present below the images.}
  \vspace{-10pt}
  \label{fig:real_experiment}
\end{figure}

In the real-world experiments, we tested two reconfigurable robots as shown in Fig.\ref{fig:real_experiment}. To validate if our model trained solely in a simulated environment could effectively predict configurations in the real world, neither robot was seen during the training of the meta-self-model. Each robot collects 10 trajectories in the real world for about 10 seconds. The state data can all be obtained through the Intel Realsense tracking camera T265. Every step involved an action and the subsequent change in the robot's state, resulting in an input size of 10x16x30. Both robot leg configurations were predicted with 100\% accuracy, demonstrating the meta-self-model trained in the simulation can also predict robot leg configurations through dynamic data in the real-world environment.

The joint prediction accuracy varied between the two robots. For the robot shown in Fig.\ref{fig:real_experiment}a (left), we achieved a prediction accuracy rate of 50.0\%. Meanwhile, for the robot in Fig.\ref{fig:real_experiment}b (right), the joint prediction accuracy was slightly higher, standing at 66.7\%. Based on limited interaction steps, the overall prediction accuracy validates the effectiveness of our method in a real-world scenario.


\section{CONCLUSIONS}

This paper presents a meta-self-modeling approach for robots to understand their own morphology from proprioceptive data. A 12-DOF reconfigurable legged robot was designed, enabling the creation of a diverse dataset of 200,000 unique robot configurations. A Multiclass-Multioutput Robot Morphology Classifier was developed to predict unseen robot configurations by observing limited self-movement data. The model architecture consists of a robot signature encoder to extract latent morphological representations from state sequences, and a configuration decoder with multiple classification heads to predict the robot's leg positions and joint angles.

Our experimental evaluations, conducted in both simulated environments and real-world settings, affirm the robustness and reliability of our meta-self-model. In the simulation, the model showcased its ability to predict leg configurations with exceptional accuracy, illustrating its deep understanding of the intricate relationship between motion dynamics and morphology. Real-world experiments further validated the model's applicability, demonstrating that it can successfully translate its predictions from simulation to physical robots. However, it is noteworthy that the model's performance in predicting joint positions encountered limitations, particularly in real-world tests. This discrepancy underscores the challenges in accurately predicting the robot morphology in different environments.

This work makes progress in robotic self-modeling by learning shared dynamics across diverse morphologies. The latent space of the model captures an understanding of how morphology relates to dynamic motion, which could enable future applications such as adaptive control, precise dynamics, and visualizing robot structures. The 12-DOF reconfigurable robot design and dataset of 200k configurations are open-sourced to enable further research. Overall, this meta-self-modeling approach offers a path toward robot autonomous identification through the proprioceptive to understand body structure and dynamics.

\bibliographystyle{IEEEtran}
\bibliography{IEEEabrv, ref}

\end{document}